\documentclass[a4paper,10pt,preprint,5p]{elsarticle}

\usepackage{blindtext}
\usepackage{lineno,hyperref}
\usepackage{bookmark}
\usepackage{amsmath}
\usepackage{amssymb}
\usepackage{amsfonts}
\usepackage{bm}
\usepackage{xcolor}
\usepackage{xfrac}
\modulolinenumbers[5]
\label{sec:label}

\begin{document}

\begin{frontmatter}
\title{Learning multiobjective rough terrain traversability}
\author[UMU]{Erik Wallin}
\author[UMU]{Viktor Wiberg}
\author[UMU]{Folke Vesterlund}
\author[SLU]{Johan Holmgren}
\author[SLU]{Henrik Persson}
\author[UMU]{Martin Servin\corref{mycorrespondingauthor}}
\cortext[mycorrespondingauthor]{Corresponding author}
\ead{martin.servin@umu.se}
\address[UMU]{Ume\aa\ University, SE-90187, Ume\aa, Sweden}
\address[SLU]{Swedish University of Agricultural Sciences, SE-90183, Ume\aa, Sweden}

\begin{abstract}
We present a method that uses high-resolution topography data of rough terrain, and ground vehicle simulation, to predict traversability. 
Traversability is expressed as three independent measures: the ability to traverse the terrain at a target speed, energy consumption, and acceleration. 
The measures are continuous and reflect different objectives for planning that go beyond binary classification.
A deep neural network is trained to predict the traversability measures from the local heightmap and target speed. 
To produce training data, we use an articulated vehicle with wheeled bogie suspensions and procedurally generated terrains.
We evaluate the model on laser-scanned forest terrains, previously unseen by the model.
The model predicts traversability with an accuracy of 90\%.
Predictions rely on features from the high-dimensional terrain data that surpass local roughness and slope relative to the heading.
Correlations show that the three traversability measures are complementary to each other. 
With an inference speed 3000 times faster than the ground truth simulation and trivially parallelizable, 
the model is well suited for traversability analysis and optimal path planning over large areas.
\end{abstract}


\end{frontmatter}


\section{Introduction}
Terrain traversability depends on the geometrical and physical properties of the terrain and the vehicle \cite{Papadakis2013,Guastella2021}.
Predicting it in advance facilitates planning and is a key component of remote and autonomous driving. Deficient or inaccurate information about traversability leads to substandard paths that consume excess fuel and time, with unnecessary risk of damaging the equipment and environment. It is common to classify areas as being either traversable or non-traversable. For vehicles designed to operate in rough terrain, binary classification, even with directional dependency, is of limited use.
On most natural terrain it yields an abundance of feasible paths with no way to distinguish between their quality.
In such cases, a continuous measure is more beneficial but lacks to consider additional aspects, such as energy consumption and mechanical wear.
A more practical approach is to treat traversability as a multivalued measure that captures how vehicle dynamics are affected by local topography.

We assume access to high-resolution terrain data through airborne laser scanning over large areas.
In Sweden, the entire country is scanned at sub meter accuracy by the Swedish Land Survey with a repeat cycle of approximately seven years.
These data capture terrain features at scales that interact with the vehicle's geometry and dynamics in a way that prevalent traversability models, developed for coarse terrain data, do not.
The availability of high-resolution surface data raises the need to develop precise models that predict vehicle traversability in rough terrain.

With a 3D multibody dynamics simulation of a vehicle driving on a virtual terrain, the interaction can be 
captured in detail.  The observed traversability is then automatically a function of the vehicle geometry, dynamics, 
and of the terrain topography.
The traversability of a terrain could in principle be systematically probed by running multiple simulations along different paths.
However, online prediction of traversability or optimal path planning on large areas
typically require simulation speeds much faster than is accessible today.
This has been addressed in \cite{Chavez-Garcia2018} by running simulations in advance of a vehicle 
traversing different terrains  while sampling the
local heightmap.  The dataset was then used for training a deep neural network 
to make predictions given the vehicle's local heightmap and heading as input.
Evaluating the model was orders of magnitude faster than the simulation.
However, the study focused on binary classification, with the classes traversable and non-traversable, 
and considered a four-wheeled skid-steer robot with no articulation and limited capabilities of traversing rough terrain.


To improve traversability analysis and planning for vehicles in rough terrain, we explore how to predict continuous measures of traversability based on multibody simulations.
%
In path planning, the flexibility to define a cost function according to preference is limited to the traversability information available. 
%
We propose that energy consumption and acceleration, alongside \emph{locomotion}, i.e. the ability to move at prescribed speed, are three complementary measures that, when combined, can yield quality paths that fit a wide variety of demands.
For learning the traversability measures, we use a dataset from an articulated vehicle driving on generated terrains with spatial resolution much finer than the size of the vehicle.
The model takes target velocity and local topography as input, processed by a neural network for feature extraction.
We suggest that these features relate vehicle dynamics to terrain topography, and has significant effect on traversability.
To unravel how learning from generated terrains transfers to predictions on scanned terrains, we compare predictions with simulation ground truth.
The comparison, along individual paths and over large regions, reveals the strengths and weaknesses of the model.
We apply the model to optimal path planning on a scanned forest terrain.

\section{Related work} 
Classical and machine learning (ML) based methods for analysing and predicting traversability are 
reviewed in \cite{Papadakis2013} and \cite{Guastella2021}, respectively.  
In the literature, \emph{traversability} is generally understood as the ability of a ground vehicle to 
move over a terrain region given some objective function and criteria for admissible states. 
Non-admissible states typically include collisions and unrecoverable states. 
The ability may be represented using discrete classes (classification) or with a continuous score (regression).  

Appearance-based methods approach traversability analysis as an image-processing problem, 
e.g., distinguishing between different types of soil and vegetation with distinct costs for traversal 
\cite{Brooks2007} or wheel slip \cite{Bouguelia2017}.  In \cite{Quann2020}, a Gaussian process regressor was used
for learning to predict the power consumption on different terrains from satellite imagery in addition to heading and slope.

Geometry-based methods first transform LiDAR, or other depth data, to a 3D representation of the terrain.
Traversability is then analyzed with respect to geometric features of the terrain \-- such as height, roughness, slope, and 
curvature \-- and to the vehicle's geometry and mechanics.  Classically, the extraction of terrain features
and the comparison with vehicle properties are two distinct computational processes.  This becomes challenging
when high-dimensional data is involved. With deep learning based methods it is natural to integrate
these processes in the convolutional kernels, which was done in the aforementioned work \cite{Chavez-Garcia2018} 
and in \cite{Zhu2020} from 3D LiDAR data and driving trajectories from experts.
In \cite{Arena2021}, a shallow neural network was trained on simulated data to predict multiple
robot-specific traversability maps, so that the best suited robot can be selected along with a path. 

Traversability is of major importance in forestry \cite{Eriksson2014} where heavy vehicles, weighing up 
to 40 tons when fully loaded, traverse rough and sometimes weak terrain. 
In \cite{suvinen2009}, digital soil maps were combined with discrete elevation maps to predict the traversability, quantified as driving resistance and divided into rolling, slope,
and obstacle resistance.  It was concluded that a resolution finer than 25 m is needed for predicting obstacle resistance. 
A method for optimization of forestry extraction routes from discrete elevation maps and depth to water maps 
is described and evaluated in \cite{Flisberg2020}.  The weighted objective functions include the time, fuel, 
soil disturbance, and additional costs for driving with different headings in sloped terrain.  The
weight factors are determined by inverse optimization using best practice solutions by experienced
forest professionals.  The study is limited to a resolution of 2~m and features below this 
length scale are not considered.  Large obstacles lead to exclusion of no-go cells or
route segments. 

\section{Method}

\begin{figure*}[ht]
  \centering
  \includegraphics[width=0.9\textwidth]{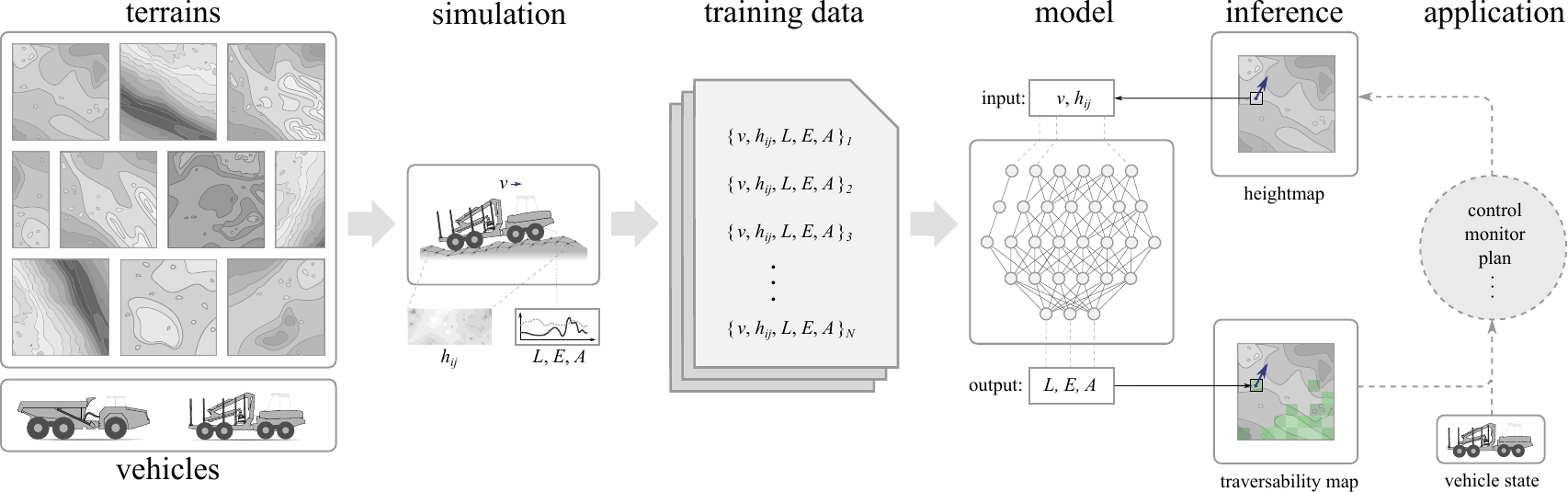}
  \caption{Illustration of the procedure for learning rough terrain traversability from simulations, for some vehicle, and applying it to previously unseen terrains.}
  \label{fig:method}
\end{figure*}

An overview of the method, from simulation modelling, to gathering data, and training the network is illustrated in Fig.~\ref{fig:method}.  


\subsection{Vehicle model}
\begin{figure}
  \centering
  \includegraphics[width=0.45\textwidth]{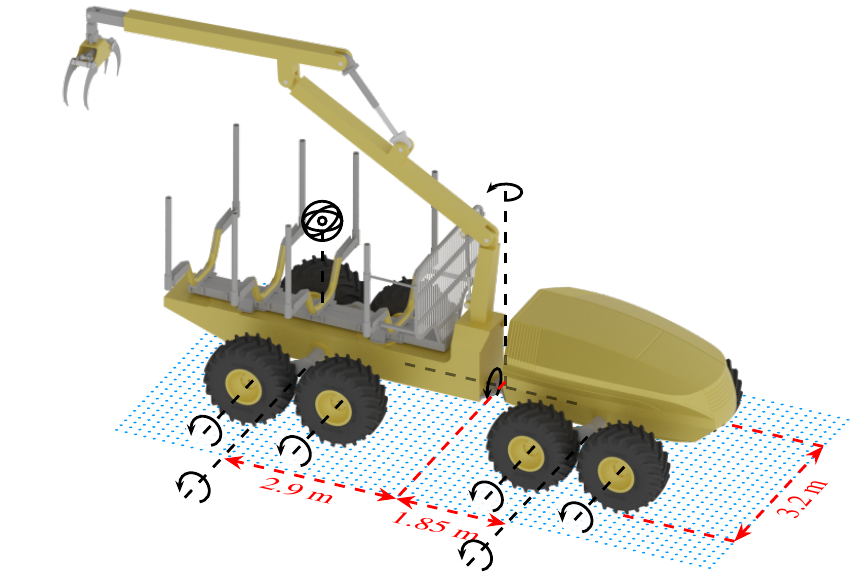}
  \caption{The 3D model of a medium-sized forwarder with primary joint axes and dimensions indicated.
  The sampling points of the local heightmap and placement of the virtual accelerometer are also shown.}
  \label{fig:forwarder}
\end{figure}

The vehicle model is that of a medium-sized, eight-wheeled, bogie type forwarder, which 
are used in forestry for transporting logs from the harvesting site to the road for transport on trucks. 
The model is shown in Fig.~\ref{fig:forwarder}. It has a front and rear frame connected by a waist articulation joint that allows relative rolling.
Each frame has a pair of bogies with two wheels each.  
Ground clearance, weight, boogie to waist distances, and tyre dimensions
are the same as for a Komatsu~845 forwarder, while being $0.6$~m wider.
The length is 9.3~m and the total weight is 16,950~kg without load.
Each wheel is driven with a hinge motor that can deliver a maximum torque of $\pm40$~kN/m
to achieve a set target angular speed $\omega = v/r$, for the wheel radius $r = 0.675$~m.
Consequently, the wheels will slip when the friction force is insufficient or if they loose contact with the ground.
In a steep slope, or another situation with large external forces, the wheel motors may be too weak to reach and hold the target speed. 
For simplicity, the tyres are given a cylindrical collision shape, although rendered with a more resolved geometry. 
In the simulations, the vehicle is assigned some target speed $v$, which is applied to all wheels.
The model accounts for frictional wheel slip, but slip from tyre and terrain deformations are not modelled explicitly, and neither is rolling resistance.
A video illustrating the vehicle kinematics is available as supplementary material in Appendix A.

\subsection{Terrains}
The terrains are represented with a global heightmap $h(x,y)$ using a regular Cartesian grid.
The heights are linearly interpolated between the grid points to define a polygon surface for computing contact points between the vehicle and the ground.

\subsubsection{Procedurally generated terrains}
\begin{figure}[ht]
   \centering
   \includegraphics[width=0.45\textwidth]{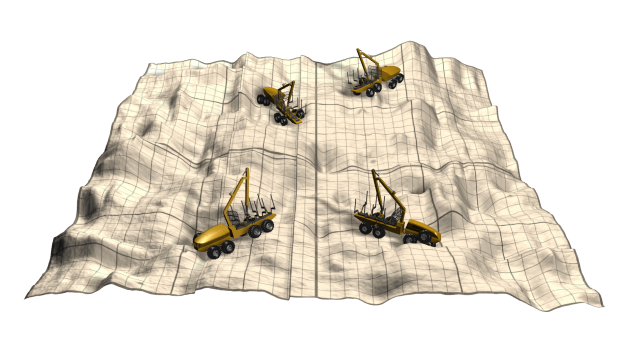}
   \includegraphics[width=0.45\textwidth]{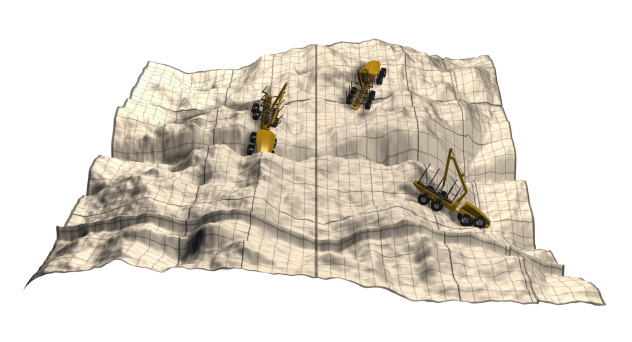}   
   \caption{Two procedurally generated terrains used for generating training data, with examples of a vehicle stuck at different positions. The side length is 50~m.}
   \label{fig:proceduralTerrains}
 \end{figure}

To have control over the terrain difficulty we use procedurally generated terrains for training.
They are made of Perlin noise \cite{perlin1985image}, Gaussian functions, step functions, and semi-ellipsoids, to represent unevenness, bumps, pits, barriers, ditches, and steps, see Fig.~\ref{fig:proceduralTerrains}.
We generate 40 terrains with a size of $50\times50$~m$^2$ and a grid resolution of $0.05$~m, where 30 of them are used for training and 10 for validation.
We found that a dataset with a relatively even distribution of good and poor traversability aids the neural network's capability to generalize to unseen terrains and predict traversability scores in the full range. 

\subsubsection{Laser scanned terrains}
For testing and demonstration we use terrain data from the SCA Laxsj{\"o} Digital Testsite in Sweden. A 600~ha subset of the totally 50,000~ha 
test site was scanned with airborne laser scanning operated from a helicopter. The system Riegl LMS-Q680i 
used a pulse repetition frequency of 400 KHz and the scanning frequency was 135 Hz. The field of view was 60 degrees, 
the nominal flight speed 20 km/h, and the altitude 70 m above ground level. The nominal swath width was 90 m and 
the nominal point density ranged from 490 points/m$^2$ to 654 points/m$^2$, with an average of 593 points/m$^2$.

\begin{figure}[ht]
  \centering
  \includegraphics[width=0.45\textwidth]{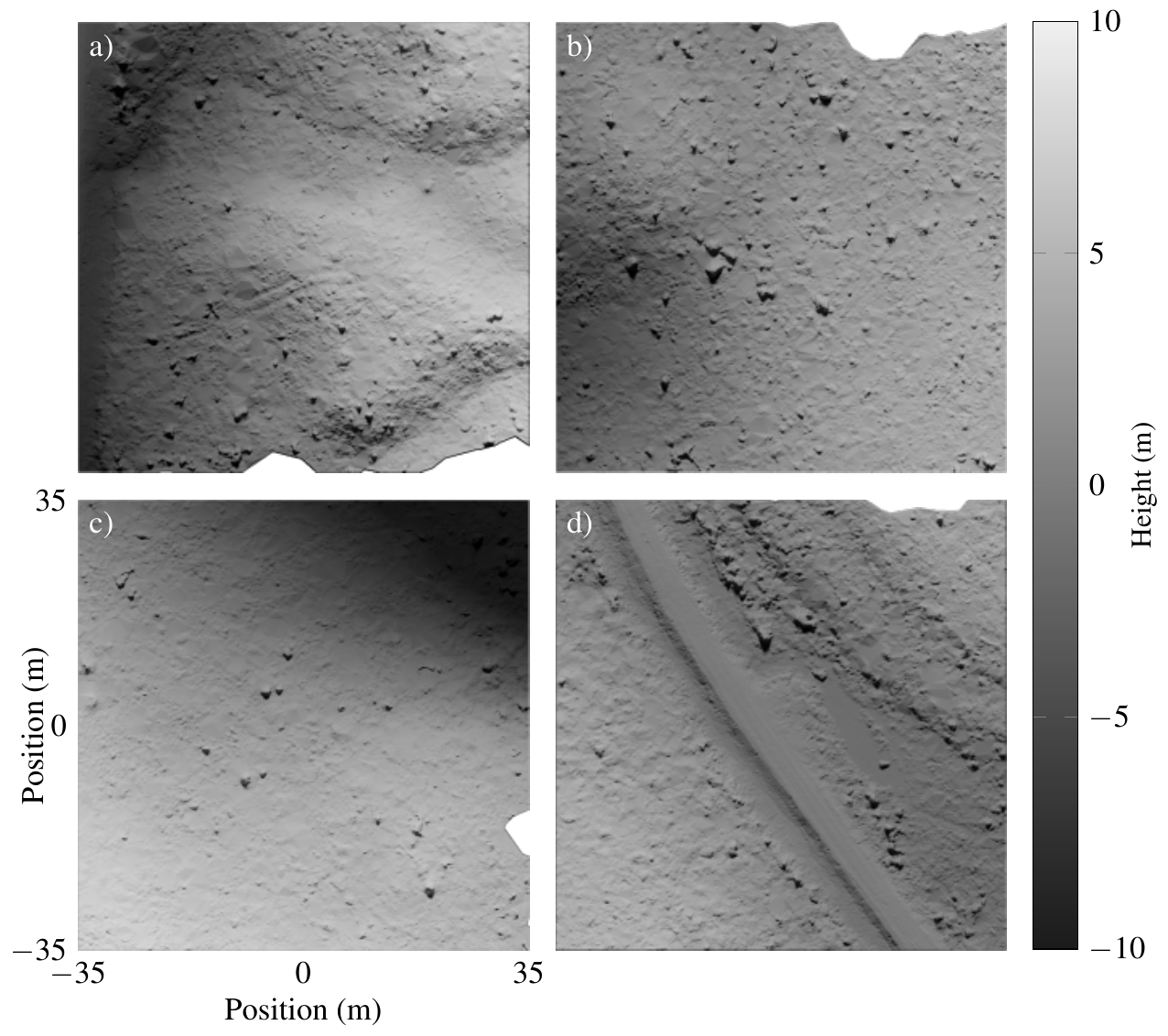}
  \caption{Rendered examples of high-resolution scanned terrains from SCA Laxsj{\"o}n Digital Testsite.
 Larger rocks are clearly visible.}
  \label{fig:scannedTerrains}
\end{figure}

The point cloud was classified into ground and vegetation points, using the algorithm~\cite{Axelsson1999} implemented 
in the software TerraScan. The ground elevation was interpolated, using a triangular irregular network, 
at the planar location $(x,y)$ of all laser returns.  The height value of each return was then replaced with the distance to the 
ground elevation.  A digital elevation map with $0.1$ m resolution was created by interpolation of ground elevation of 
the raster cell locations.  In addition, a raster with the same resolution was created by setting a cell value to the height value 
of the lowest laser return located within the raster cell. This raster was input to an active contour algorithm~\cite{elmqvist2001}. 
The algorithm moves an elastic surface from below ground level and upwards until the surface is attached to continuous objects near the 
ground, e.g. stones. In the next step, surface heights with stones were added to the raster cell values of the elevation map to create a model with both ground and stones included.
Samples of the high-resolution scanned terrains are found in Fig.~\ref{fig:scannedTerrains}.
    
\subsection{Simulations and data collection}
We simulate terrains, vehicles, and their interactions through contacting rigid multibody dynamics at simulation frequency 60~Hz using the physics engine AGX Dynamics \cite{AGX2021}. 
To collect data for training and validation we import a procedurally generated terrain and initialize a vehicle at random position and heading, letting it drop to the surface and relax.
After relaxation, data collection starts and the vehicle attempts to accelerate to its target speed $v \in [0.07, 2.69]$~m/s. 
Data is collected at 20~Hz by sampling the local heightmap around the vehicle and recording observations.
The local height map, centred at the articulation joint, is represented by a $10 \times 5$~m$^2$ area with $64 \times 32$ gridpoints, which follows the vehicle and its heading, see Fig.~\ref{fig:forwarder}.
Observations over the past $\tau = 1$~s are kept in a buffer, from which measures of traversability are calculated.
The vehicle is respawned if it is unable to stabilize at the start, reaches the boundary of the terrain, or gets stuck.
A vehicle can get stuck by the chassis hanging on the terrain, preventing one or several tyres reaching the ground for sufficient traction, by wheel slip, or if the motors do not provide sufficient torque for climbing a steep slope or obstacle. The vehicle can also overturn.
Example simulations are shown in a supplementary video in Appendix A.

To speed up the data generation, multiple vehicles are simulated simultaneously on the same terrain, with mutual collisions disabled.
For the training and validation data, a total of 20 vehicles drive for 500~s on each of the 30 training and 10 validation terrains.
For the generalization data, a total of 10 vehicles drive for 500~s each on 11 laser scanned terrains.

\subsubsection{Locomotion}
\begin{figure}
  \centering
  \includegraphics[width=0.45\textwidth]{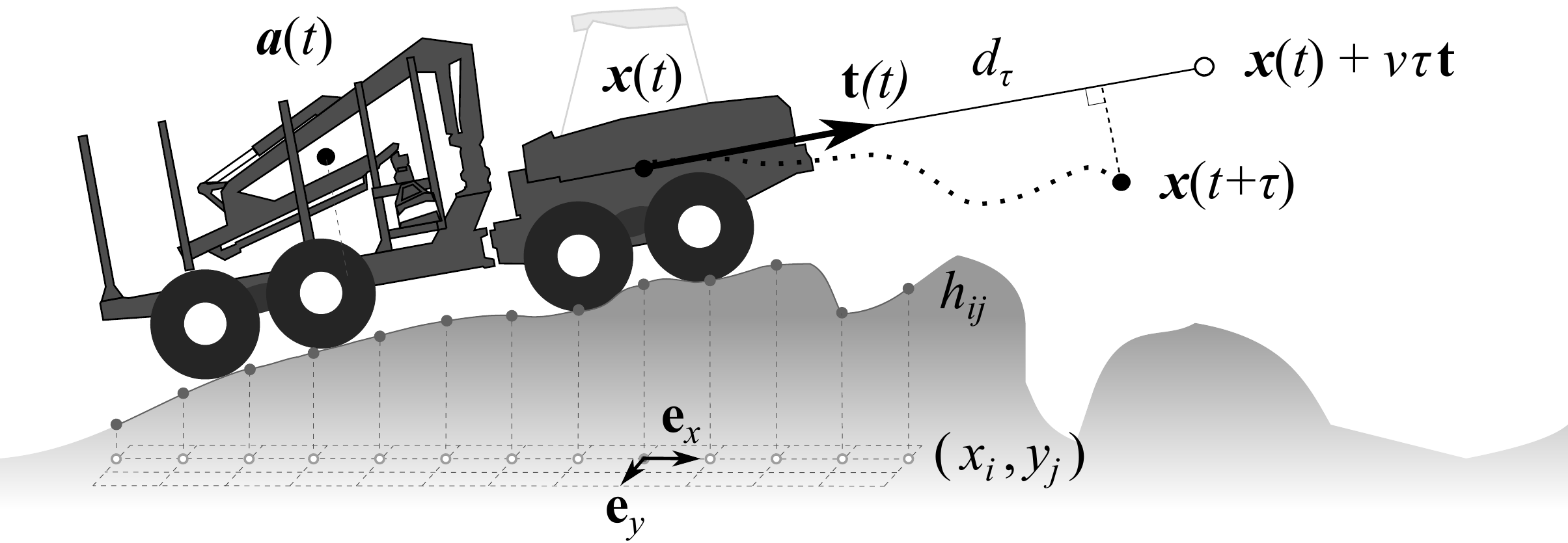}
  \caption{The locomotion measure compares the travelled distance $d_\tau$ over a time window $\tau$ 
  with the expected one given a set target speed. Acceleration is measured at a selected point
  in the rear frame.}
  \label{fig:forwarder_trafficability}
\end{figure}

We define \emph{locomotion}, $L \in [0,1]$, in terms of how close the travelled distance, 
$d_\tau$, is to the nominal distance, $d = v \tau$, for a time window of size $\tau$ and given target speed $v$.
The measure is
\begin{equation}
  L = \exp\left[ -\frac{1}{2\sigma^2} \left( \frac{d - d_\tau}{d} \right)^2 \right],
\end{equation}
with Gaussian width $\sigma=1/3$. 
The travelled distance is computed as $d_\tau \equiv \left[\bm{x}(t+\tau) - \bm{x}(t) \right] \cdot \bm{t}(t)$ using the
vehicle's heading, $\bm{t}(t) = \bm{v}(t)/\left| \bm{v}(t) \right|$, at the start of the observation window, as illustrated in Fig.~\ref{fig:forwarder_trafficability}.
If $d_\tau < 0.2 d$ for 5 consecutive observations, the vehicle is defined as stuck and is respawned.

\subsubsection{Energy consumption}
The normalized energy consumption, per unit travelled length, is computed from the work exerted by the wheel motors over the time window $\tau$ as
\begin{equation}
  E = \frac{1}{d_\tau E_0} \int_t^{t+\tau} c\sum_i P_i(t) \mathrm{d}t,
\end{equation}
where $P_i(t) = \omega_i(t) M_i(t)$ is the power exerted by each motor $i$ running with angular speed $\omega_i$ and torque $M_i$.
The coefficient $c$ is the efficiency of the motors, which is simply set to $c=1$ in the present paper.
$E_0 = 700$~kJ/m is used for normalization, this being five times larger than for driving up a 45$^\circ$ incline.
The energy consumption is clipped to the range $[0, 1]$, and for negative travelled distances, $d_{\tau} < 0$, it is set to $1$.



\subsubsection{Acceleration}
Acceleration is associated with mechanical stresses on the vehicle construction, risk for hazardous load displacements, and is harmful and uncomfortable to any human riding with the vehicle.
Therefore, a virtual accelerometer is placed $1.5$~m above and $0.5$~m behind the center of gravity of the rear frame, see Fig.~\ref{fig:forwarder_trafficability}.
We use the normalized peak acceleration during the time window
\begin{equation}
  A = \max_{\bar{t}\in[t,t+\tau]} \left[ a(\bar{t}) \right]/A_0
\end{equation}
as measure for acceleration, where $A_0=100$~m/s$^2$, and the acceleration is clipped to the range $[0, 1]$.

\subsection{Model training and architecture}
A deep neural network is trained to predict the traversability measures $[L,E,A]$ from a local heightmap and target speed $v$.
To simplify learning we normalize the velocity input and offset the heights such that the array midpoint is at height 0.

\begin{figure}[ht!]
  \centering
  \includegraphics[width=0.5\textwidth]
      {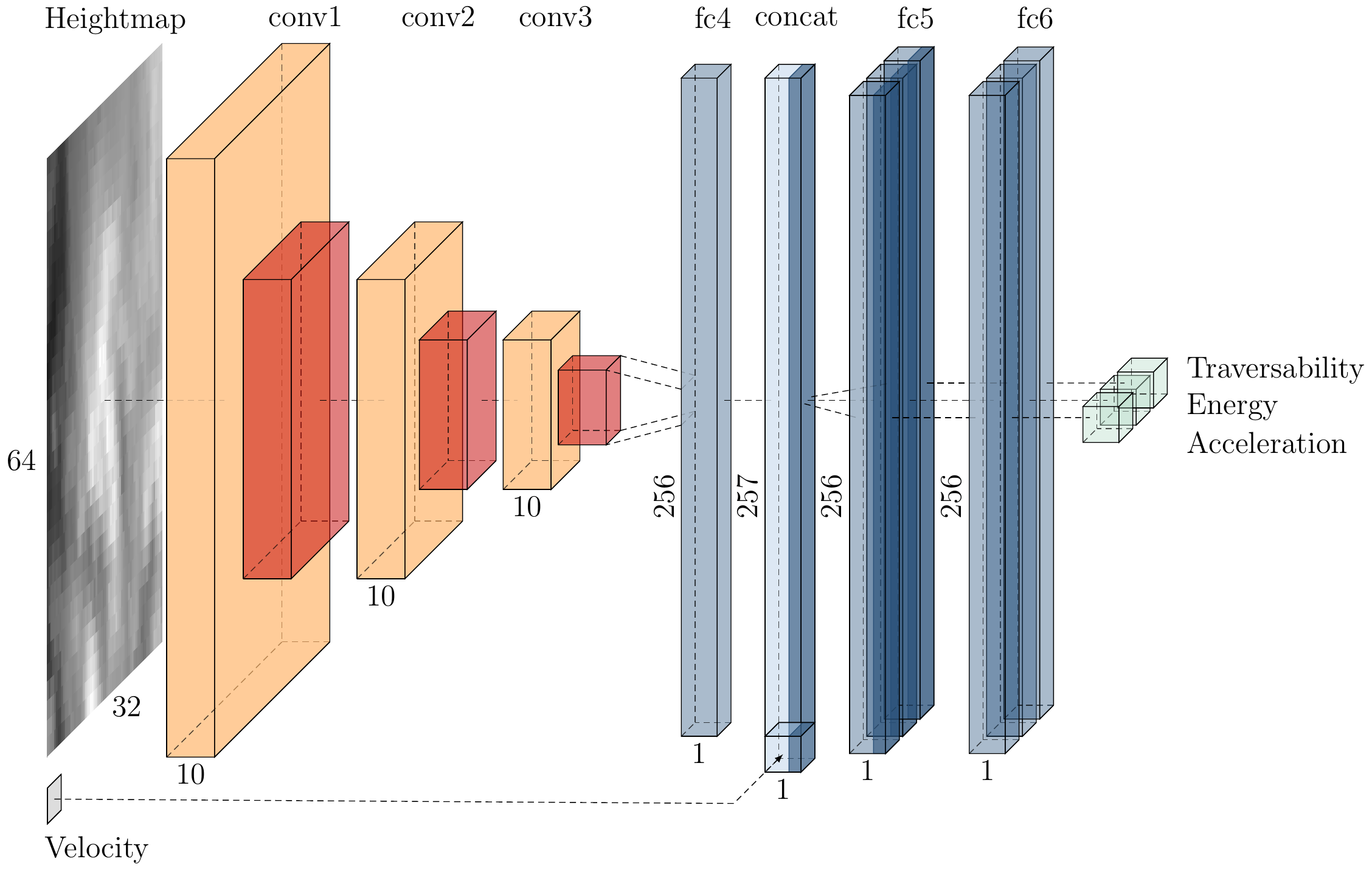}
\caption{Illustration of the neural network architecture.}
  \label{fig:model}
\end{figure}

The network consists of two input branches and three output branches, which extends the architecture of \cite{Chavez-Garcia2018} to output multiple traversability measures and include the target velocity as input, see Fig.~\ref{fig:model}.
To extract local terrain features, the input heightmap is passed through three convolutional layers.
Each layer contains 10 filters of size $3\times3$, followed by max pooling with window size $2\times2$ and the same stride.
After the convolutions, the output is flattened and passed to a fully connected layer with 256 nodes.
The heightmap features are then concatenated with the velocity input and split into three separate branches.
Each branch consist of two sets of 0.1 rate dropout and fully connected layers with 256 nodes.
All layer use ReLU activations, except for the final fully connected layer that has a single output node and linear activation.

Hyperparameter tuning was performed by varying the number of convolutional filters, the number of nodes in the fully connected layers, the dropout rate, if to use batch normalization, as well as the learning rate of the Adam optimizer, batch size and shuffle size.
We use mean absolute error as loss, where each of the three traversability measures contributes equally, and pick the model with the lowest validation loss.

\section{Results}

The trained model has a \emph{generalization loss} of $7.7\%$ that comes from evaluating the model on data from a set of 11 scanned terrains, unseen during training and testing, see Table~\ref{table:model_performance}. This is the type of terrains where the model will actually be applied.
Variations in loss and the relatively few terrains explains why the acceleration has a lower validation than training loss.
Although the generalization loss is a useful metric to quantify model performance, it provides limited insight into the strengths and weaknesses of the model.



\begin{table}[ht!]
  \centering
  \begin{tabular}{lcccc} 
    \hline
                  & loss  & $L$ & $E$ & $A$ \\
    \hline
    Training error (\%)        & $10.3$ & $12.3$ & $10.0$ & $8.4$  \\ 
    Validation error (\%)      & $10.8$ & $13.7$ & $11.1$ & $7.5$ \\ 
    Generalization error (\%)  & $7.7$ & $9.8$ & $5.4$ & $7.8$ \\ 
    \hline
  \end{tabular}
  \caption{Model performance on the training (3.7 million observation), validation data (1.4 million observations), and generalization to laser scanned terrains (0.9 million observations).}
  \label{table:model_performance}
\end{table}

\subsection{Traversability maps}
\begin{figure}
  \centering
  \includegraphics[width=1.0\columnwidth]{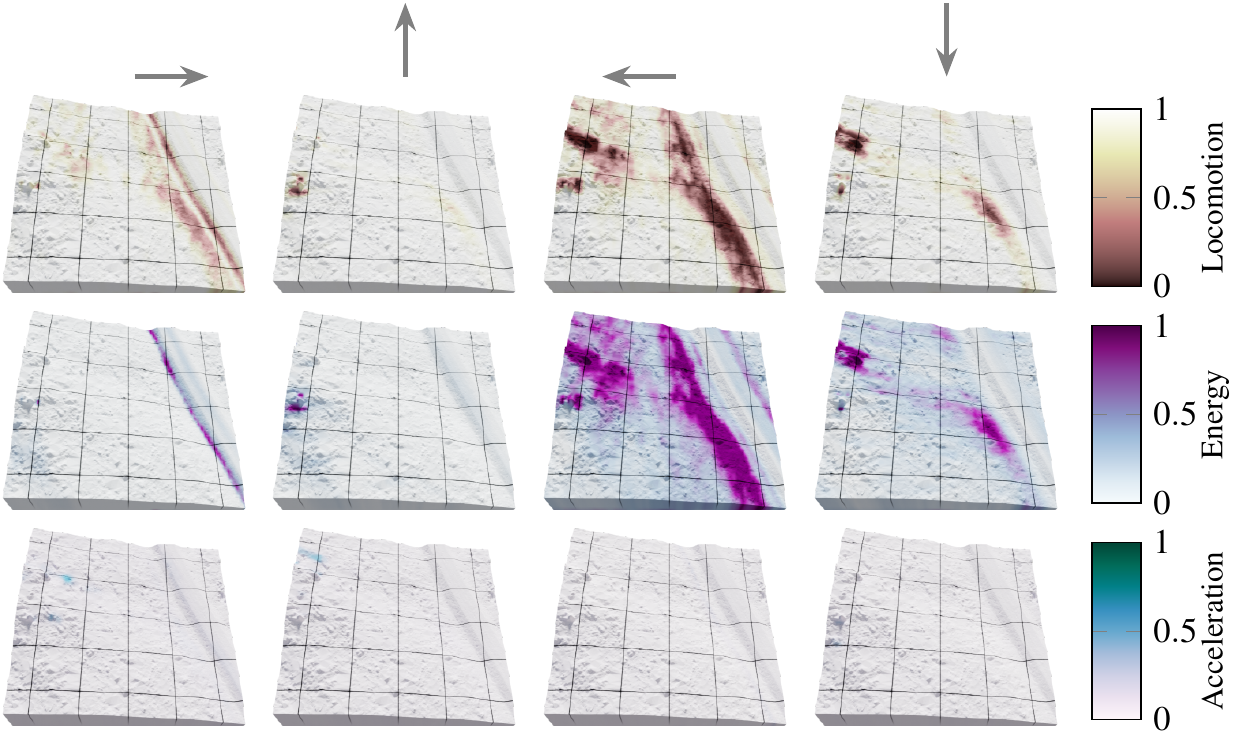}
  \caption{Traversability maps from evaluating the model on a specific scanned terrain. The size is $50\times50$~m$^2$ and the driving direction is indicated with an arrow.
  A supplementary video is available in Appendix A.}
  \label{fig:trafficability_laxsjon_4_dir}
\end{figure}

To analyze predictions, the model is swept over terrain maps in selected headings and target speeds, resulting in \emph{traversability maps}.
Sample maps from one of the scanned terrains can be found in Fig.~\ref{fig:trafficability_laxsjon_4_dir}, given a target velocity of $1.3$ m/s.
We observe that the traversability measures are highly directionally dependent and correlated with slope and local roughness. 
The road segment on the right is fully traversable in all directions, but the ditch parallel to the road can only be traversed at an angle and not head on.

\begin{figure}[ht!]
  \centering
  \includegraphics[width=0.48\textwidth]{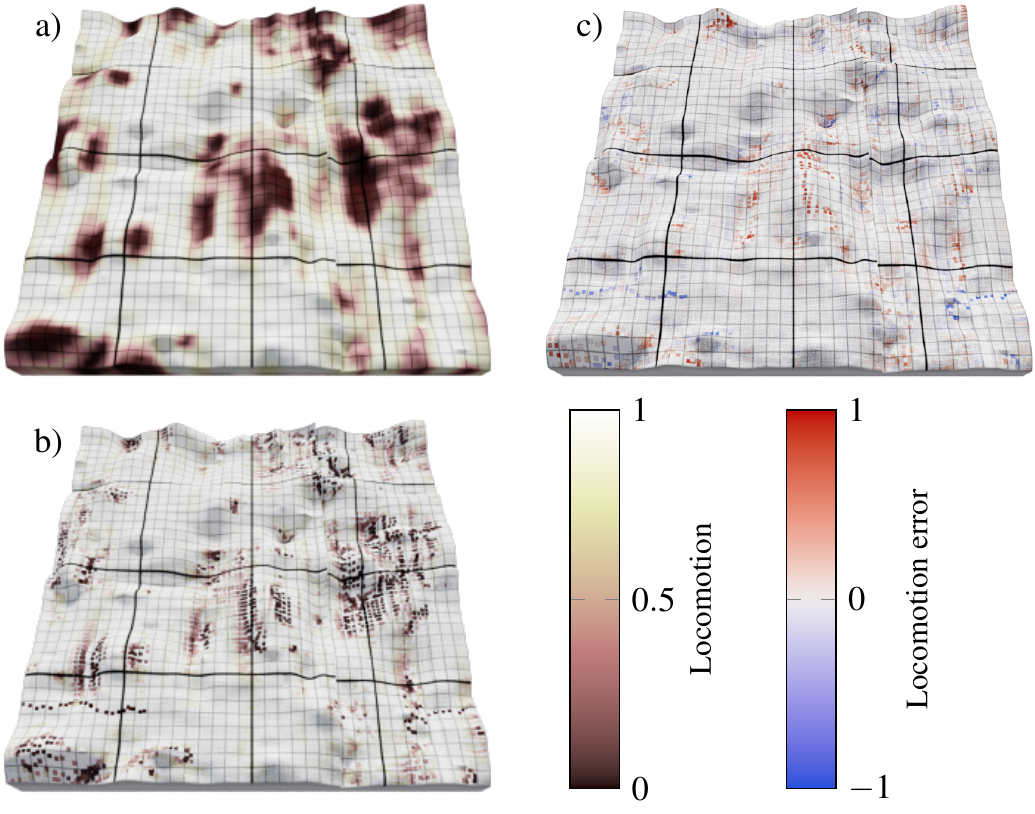}
  \caption{Model predicted locomotion map in a 3D view (a) compared to a locomotion scatter plot (b), constructed by spawning the vehicle at $100\times100$ locations on a validation terrain.
The difference between these measures and the model predictions (c), with an average error of $13.5\%$. The size of all maps is $36\times36$~m$^2$ and the driving direction is to the right.}
  \label{fig:error_procedurally}
\end{figure}

\begin{figure}[ht!]
  \centering
  \includegraphics[width=0.99\columnwidth]{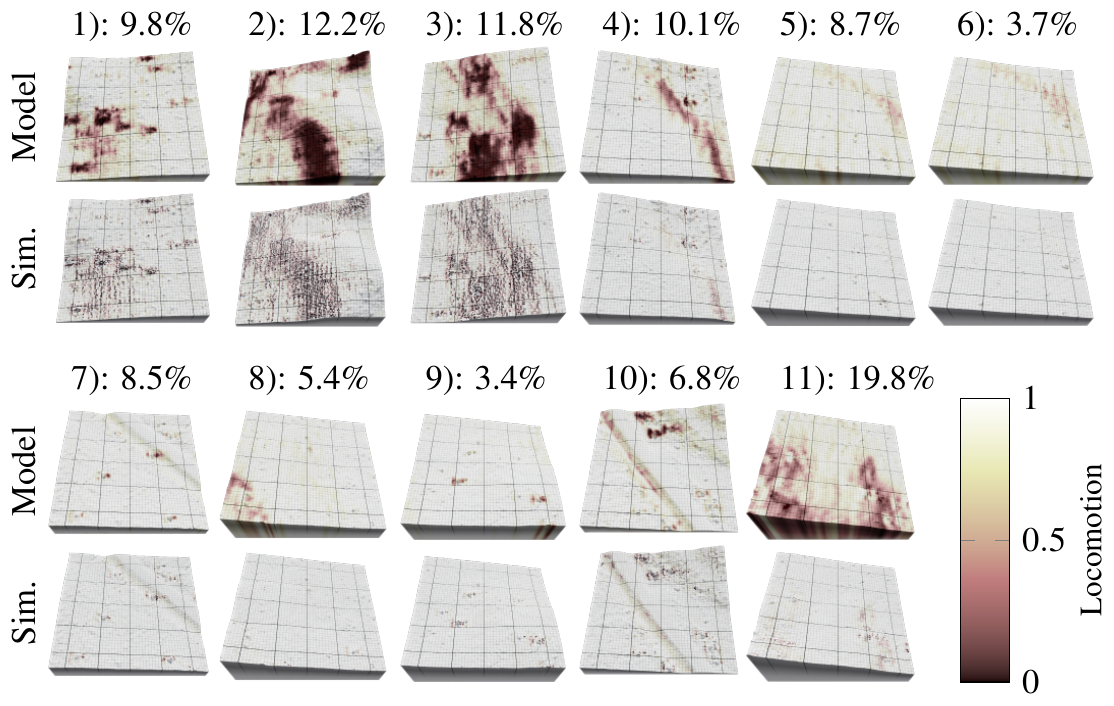}
  \caption{Model predicted locomotion map on 11 scanned terrains shown in a 3D view (first and third rows), compared to locomotion scatter plots, constructed by spawning the vehicle at $100\times100$ location.
The locomotion error is stated for each terrain.
The average locomotion error over all 11 validation terrains is $9.1\%$, with the average energy and acceleration errors being $3.1\%$ and $8.4\%$ respectively. The size of all maps is $50\times50$~m$^2$ and the driving direction is to the right.}
  \label{fig:error_laser}
\end{figure}

To compare traversability maps with ground truth simulations, we generate a dataset where we spawn the vehicle in a grid structure for given directions and velocities, i.e. similar to sweeping the model but roughly $3000$ times slower.
To place the vehicle close to the desired positions we attach it to an anchor, constrained to vertical movement with fixed yaw angle.
For each spawn position, the vehicle is dropped and relaxed, after which it is run for one observation window.
The locomotion and its error distribution for a representative procedurally generated terrain and a selected driving direction, are shown in Fig.~\ref{fig:error_procedurally}.
The average locomotion error over all validation terrains is $13.0\%$, while the average energy and acceleration errors are $12.4\%$ and $12.0\%$, respectively.
Similarly, the locomotion predicted by the model, and the simulation outcome for the 11 scanned terrains, is shown in Fig.~\ref{fig:error_laser}.
A comparison between predictions and simulation outcomes reveals a general agreement, but with a systematic difference in steep downhill sections. This is most notably seen for terrain 11, with regions of low locomotion for the model but not for the simulations.
This can be due to the model overestimating locomotion in steep downhill terrain.
However, a partial explanation is that the vehicle only has a 1~s time window to build momentum before respawning.
The relatively short observation window prevents a drop in locomotion caused by overspeeding that could occur if given the time.

For application purposes it is interesting to know how fast the model is to evaluate and how this compares to the original simulations.
The measured inference speed is 0.55 ms, taking 44~s to sweep $100\times 100$ points on a grid in 8 directions on an Intel(R) Core(TM) i7-8700 CPU @3.20GHz.
This is three orders in magnitude faster than running the simulation 
model to evaluate traversability at 1 m spacing.  As an example we note that traversability maps with 1 m resolution over the 200,000 ha
of Swedish forest terrain that is harvested each year takes 2500 CPU hours to compute, and a fraction of that in wall clock time on a powerful cluster. 

\subsection{Model performance along example paths}
\begin{figure}[ht!]
\centering
  \includegraphics[width=0.3\textwidth]{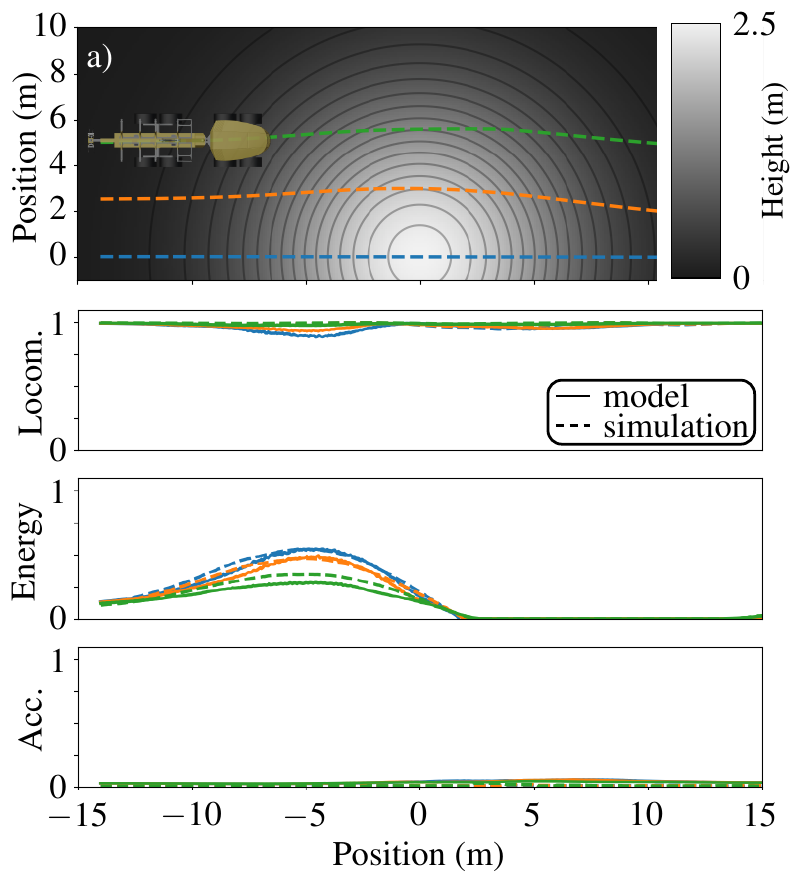}\\
  \vspace{5mm}
  \includegraphics[width=0.3\textwidth]{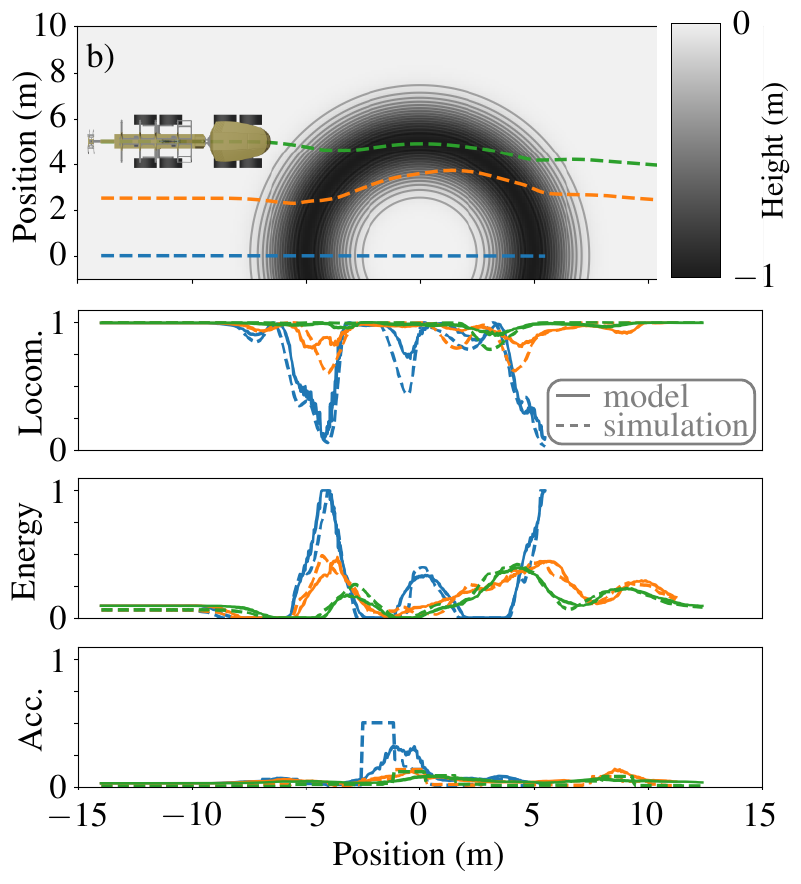}\\
  \vspace{5mm}
  \includegraphics[width=0.3\textwidth]{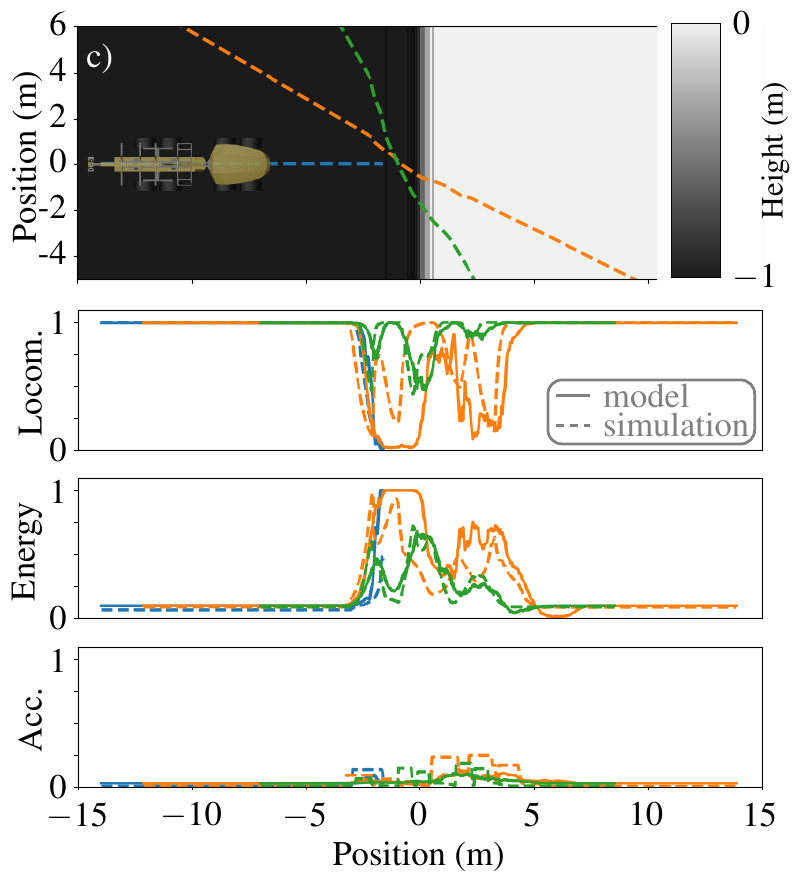}
  \caption{Comparison between simulated and model predicted traversability measures for three selected trajectories in green, orange, and blue, over a Gaussian hill (a), a circular ditch (b), and a step (c).}
\label{fig:artificial_tests}
\end{figure}

To see in detail how the model predictions compare to the simulation ground truth, we simulate the vehicle driving on three specific synthetic terrains.
The synthetic terrains are designed to be sensitive in terms of success or failure with respect to vehicle dynamics. They allow us to study isolated behaviours with intuitive outcomes, which cannot be untangled in more complex situations.
The results on a Gaussian hill (2.5 m high and 12 m wide at half maximum), a circular ditch (1 m deep, 2.3 m wide at half maximum, 5 m radius), and a step (1 m tall, 0.5 m cubic smoothing) are shown Fig.~\ref{fig:artificial_tests}.
The general agreement is good, although we note that acceleration is generally underestimated.
The definition of acceleration with a $\max$ function introduces discontinuities, which are difficult for the model to mimic.
The model successfully captures that the ditch and step are not traversable head-on, but with considerable struggle, at an offset or angle.



\begin{figure}[ht!]
  \centering
  \includegraphics[width=0.49\columnwidth]{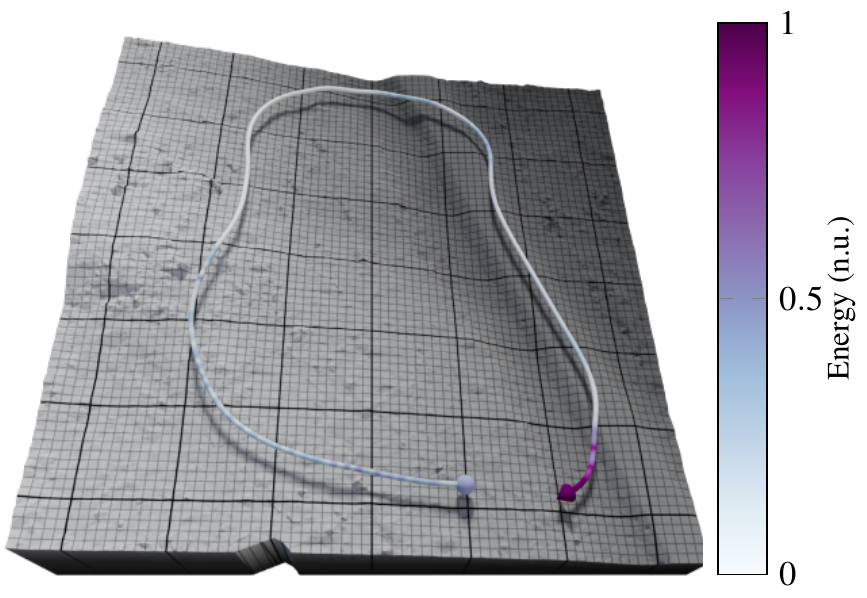}
  \includegraphics[width=0.49\columnwidth]{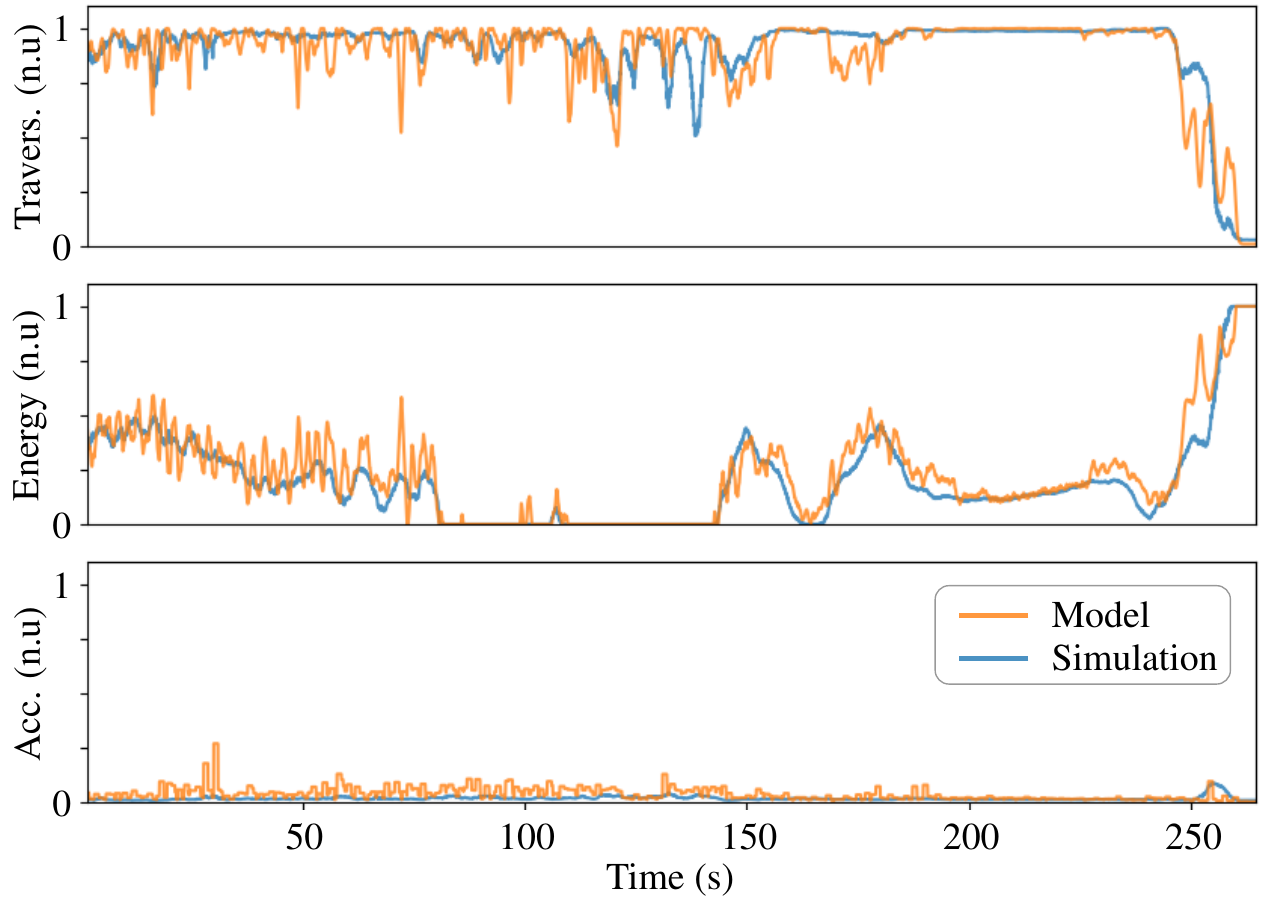}
\caption{Left shows a manually driven route on a scanned terrain (cf. Fig.~\ref{fig:trafficability_laxsjon_4_dir}) with size 70$\times$70~m$^2$. Right shows a comparison between simulated and model predicted locomotion, energy consumption, and acceleration along the route.
}
  \label{fig:results:man-path-energy}
\end{figure}

Next, we test the model in a scenario where the vehicle is turning despite only being trained on data gathered from driving straight.
We manually drive the vehicle on a scanned terrain along a curved path using articulated steering and a target velocity of 0.675~m/s while logging L, E, and A.
The same path is then swept by the model, see Fig.~\ref{fig:results:man-path-energy}.
The route starts slightly uphill on a bumpy area, well reflected by the initial energy consumption.
The descent requires minimal energy until the vehicle reaches a forest road.
At the end of the route, the vehicle leaves the forest road and encounters a non-traversable section, which results in vanishing locomotion and maximum energy consumption.
Overall, the model predicted locomotion and energy consumption agrees well with the simulation ground truth, even when the vehicle is taking
moderately sharp turns.

\subsection{Dependency on spatial resolution}
The use of traversability maps in applications requires knowing the sensitivity to the spatial resolution, as the resolution depends on how field data is collected.
To investigate the sensitivity we first resample the 11 scanned terrains to new resolutions, ranging from the original $0.1$~m up to $50$~m.
We then compare the result of sweeping the model on $100\times100$ points and 8 directions between the original and the resampled terrains.
The error is taken as the average of the pointwise difference of each traversability measure, normalized using the values from the coarsest resolution.
This resolution corresponds to removing all spatial information and evaluating the model on a flat grid and serves as the worst case prediction.

\begin{figure}[ht!]
  \centering
  \includegraphics[width=0.48\textwidth]{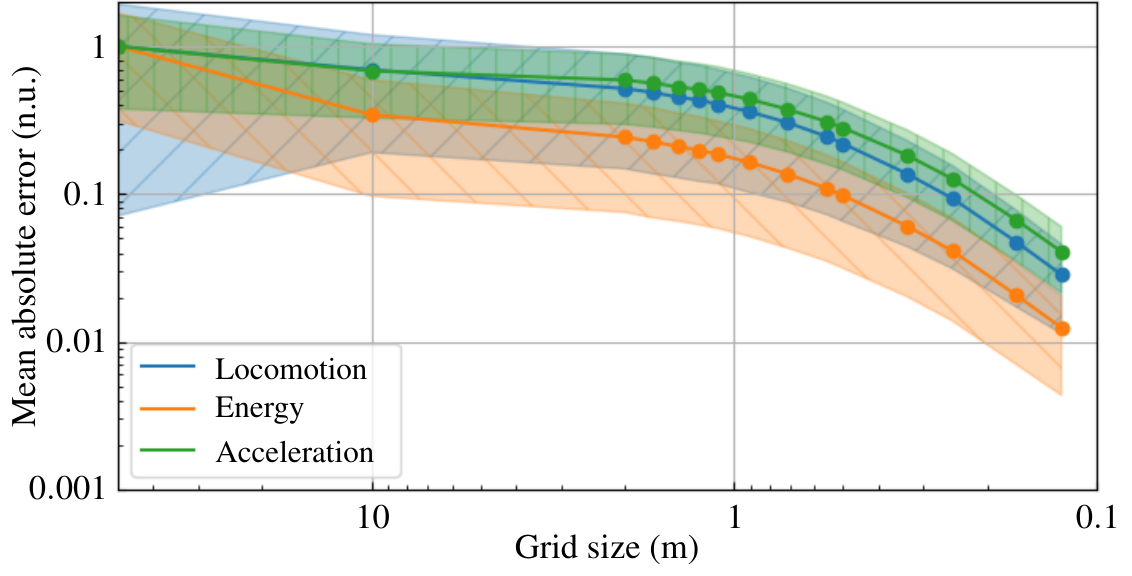}
  \caption{
The dependency of heightmap resolution on scanned terrains. The error is the pointwise average between the original resolution and resampled coarser versions, normalized by the coarsest sample, i.e. evaluating the model on a flat surface. The original resolution of 0.1~m yields 0 mean error and is omitted from the plot.
}

  \label{fig:error_resolution}
\end{figure}

The results show that locomotion and acceleration require higher resolution than the energy, see Fig.~\ref{fig:error_resolution}.
It is reasonable that these depend more on fine features, while the energy relies more on coarser features such as the slope.
We note that an error below 10\% requires at least $0.25$~m resolution.

\subsection{Velocity dependency}
\begin{figure}[ht!]
  \centering
  \includegraphics[width=0.9\columnwidth]{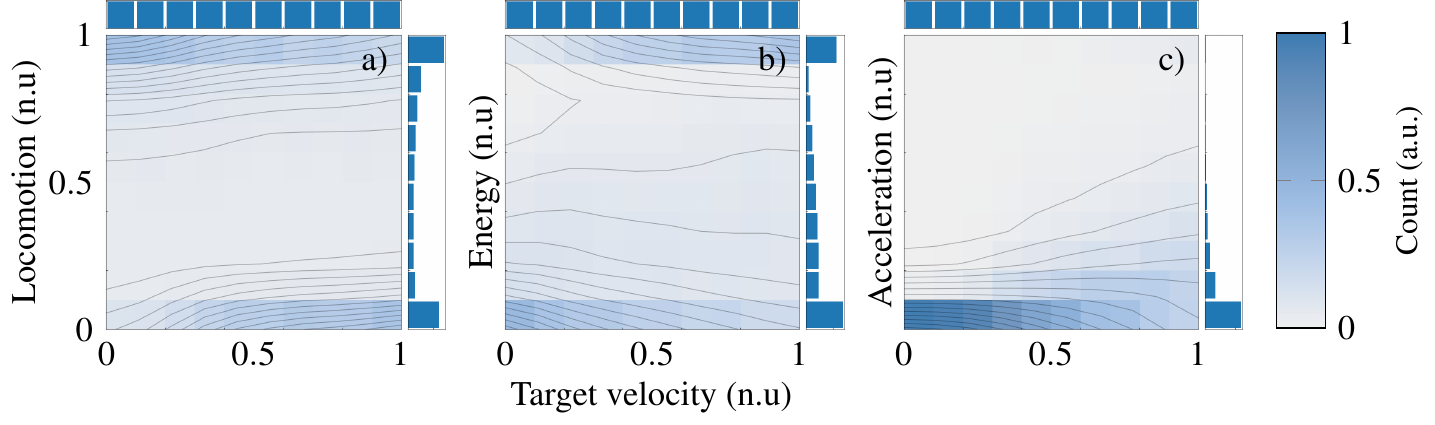}
  \caption{The dependency of velocity on locomotion (a), energy consumption (b), and acceleration (c) on the validation dataset.
The 2D histograms have been normalized column wise to compensate for the number of samples within each target velocity range.
Marginal distributions are shown on top and to the right.
}
  \label{fig:velocity_dependency}
\end{figure}

To see if the model captures the system dynamics related to the input target velocity we look at its correlation with the traversability measures. Velocity has a notable effect on the acceleration, with higher acceleration for increasing velocities, see Fig.~\ref{fig:velocity_dependency}.
For locomotion and energy the dependency is not as clear, but tends to lower locomotion and higher energy for increasing velocities.
Part of this is attributed to the data generation. A fixed amount of data is saved when the vehicle is stuck, a state more prevalent for a vehicle moving with higher speed than a slower one.

\subsection{Feature sensitivity}
We test how sensitive the trained model is to features in the local heightmap that cannot be captured by local roughness and slope relative to the heading.
Because the vehicle is left-right symmetric there are two headings, $\bm{t}$ and $\bm{t}'$ with equivalent slope in the driving direction, on an inclined plane.
Depending on local terrain irregularities, the locomotion in $\bm{t}$ and $\bm{t}'$ could be significantly different.
However, a model that depends only on slope and local roughness would yield identical results, as local roughness is independent of direction.

To test if the model captures directional dependency not explained by roughness and slope we evaluate the difference in locomotion $\mathcal{\varepsilon}^n_{L} = L(\bm{x}_n,\bm{t}_n) - L(\bm{x}_n,\bm{t}'_n)$ on a random selection of points $\bm{x}_n$ on the entire set of laser scanned terrains. The directions $\bm{t}'_n$ are found using
normals from a height field that we smooth with a Gaussian filter with standard deviation $\sigma=3$~m, based on the vehicle wheelbase. The purpose of the smoothing is that the original normal at a single point can be a poor representation of the local slope.
For each point, local roughness is evaluated surrounding a radius of 3~m as the ratio between the actual surface area and the flat surface tilted by the mean slope.
A model that is sensitive to features in the local height map beyond roughness should systematically produce nonzero $\mathcal{\varepsilon}^n_{L}$, except when the local heightmap is a plane. For such a model, we expect $\mathcal{\varepsilon}^n_{L}$ to increase with roughness, as it reflects the amount of surface irregularities surrounding the vehicle.

\begin{figure}[ht!]
  \centering
  \includegraphics[width=0.48\textwidth]{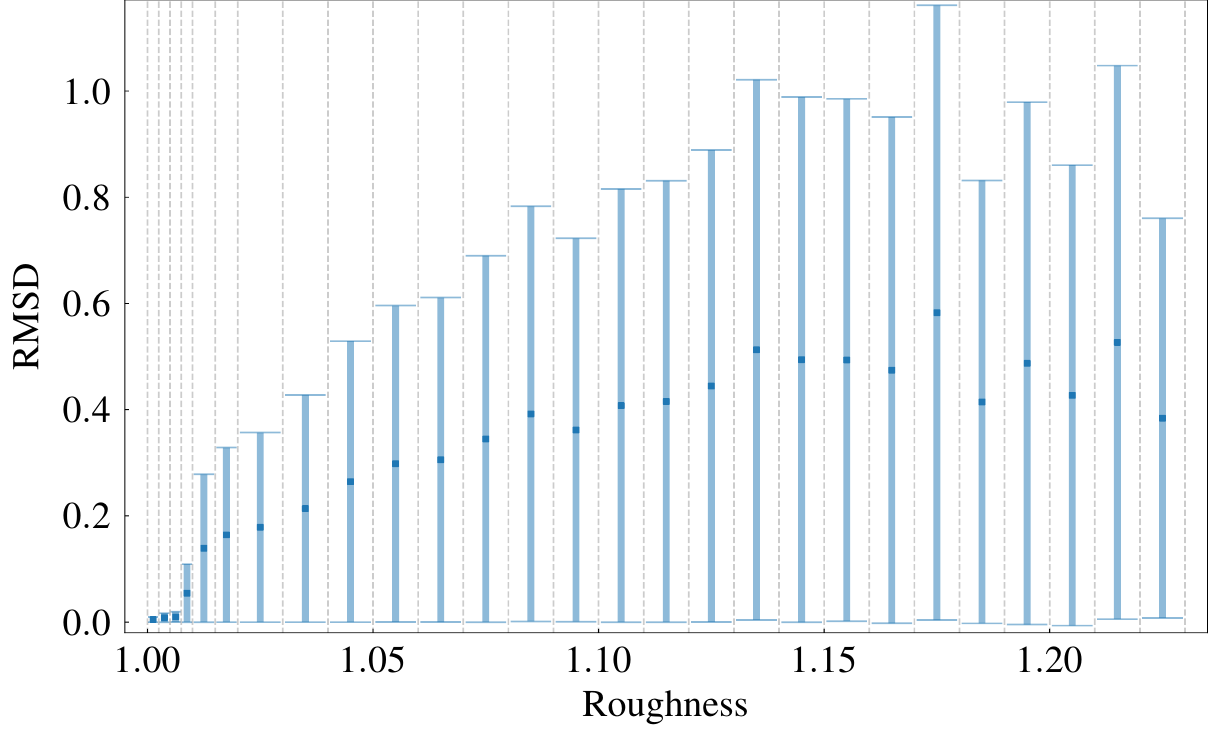}
  \caption{Root mean squared deviation and standard deviation of predicted locomotion between equivalent driving directions.
  Roughness is subdivided into four intervals from 1 to 1.01 and two intervals from 1.01 to 1.02.
  The remaining is evenly split with 0.01 spacing, where we omit intervals that contain less than 20 error measurements.
  The data is generated from 1000 randomly chosen points and headings on 32 patches of laser scanned terrains, including the 11 seen in Fig.~\ref{fig:error_laser}, resulting in a total of 32000 points.}
  \label{fig:feature_sensitivity}
\end{figure}

The resulting root mean squared deviations (RMSD) have a general trend, that as roughness increases so does the difference in locomotion, see Fig.~\ref{fig:feature_sensitivity}.
This trend together with the generalization MAE for locomotion (0.10) shows that the model not only successfully predicts locomotion, but also connects driving direction to features in the terrain topography.
As a sanity check, we see that the deviation goes to zero as roughness approaches 1, at which point the terrain is basically flat with incline.
On the other end, the plateau in RMSD for roughness greater than 1.15 is attributed to a larger portion of locomotion predictions being in the lower range.
We suggest that the distinct increase at roughness around 1.01 occurs as the model begins to take surface irregularities into consideration for predictions.

\subsection{Complementarity of the traversability measures}
\begin{figure}[ht!]
  \centering
  \includegraphics[width=0.9\columnwidth]{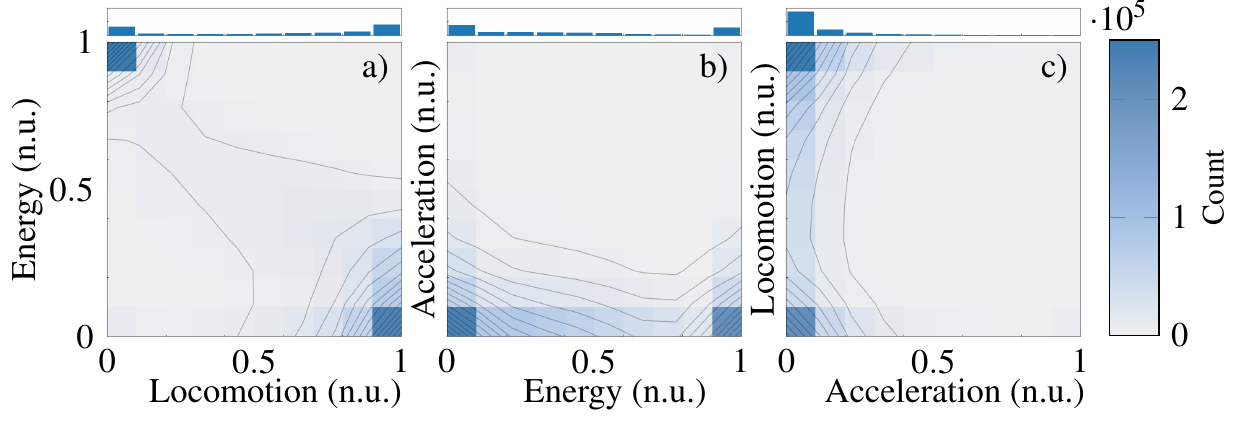}
  \caption{2D histograms showing the interdependency of locomotion, energy consumption, and acceleration on the validation dataset, with marginal distributions on top.}
  \label{fig:correlations}
\end{figure}

To check the complementarity of the traversability measures we evaluate their correlations based on the data from the validation terrains, see Fig.~\ref{fig:correlations}.
We calculate the Spearman's rank correlation coefficient $\rho_\text{corr}$, which tests for a monotonic, possibly nonlinear, relation.
As can be expected, locomotion and energy consumption are negatively correlated ($\rho_\text{corr} = -0.77$).
Points with high energy consumption are strongly correlated with low locomotion and can be explained by situations when the vehicle gets stuck with spinning wheels, or at a slope too steep to traverse.
The opposite correlation, with high locomotion and low energy consumption, is weaker.
This is natural, considering that a smooth moderate incline can be traversed at target speed but at sizeable energy requirement. Also, driving downhill requires no work, but the locomotion will drop in sufficiently steep slopes because of frictional slip.
In the intermediate regime, $L,E\in [0.1,0.9] $, the locomotion and energy consumption show only a weak negative correlation.
There are no obvious correlations between acceleration and locomotion ($\rho_\text{corr} =-0.07$) or acceleration and energy ($\rho_\text{corr} =-0.01$).
Altogether this confirms the assumption that the three traversability measures are mutually complementary.

\subsection{Planning with a multiobjective traversability model}
\begin{figure}[hbt!]
  \centering
  \includegraphics[width=0.48\textwidth]{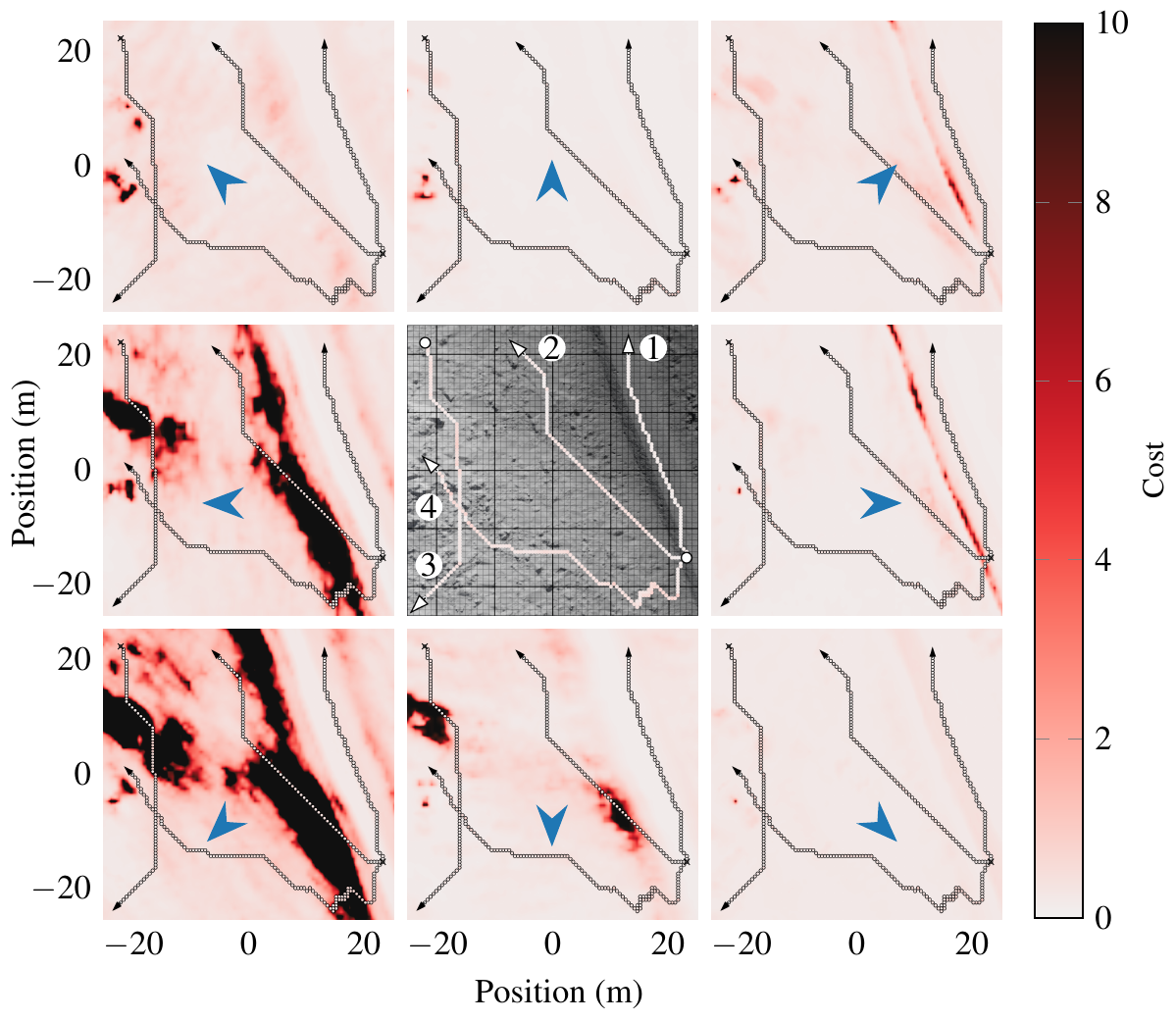}
  \caption{Maps of weighted traversability cost in eight different headings (blue arrowheads) for a scanned terrain. Four optimal paths are demonstrated.
}
  \label{fig:route_optimization}
\end{figure}

To test the use of the learned traversability model for optimal path planning with multiple objectives, we combine the
three measures into a scalar traversability cost $c_i$ and use the Dijkstra algorithm to compute the path $\{ \bm{x}_i \}_1^N$ that minimizes the accumulated cost
\begin{equation}
  \label{eq:minimization}
  C = \text{min}  \sum_{i=1}^{N} \left( \tilde{E}_i / \tilde{L}_i + A_i \right) u_i.
\end{equation}
Here $\tilde{E}_i$ and $\tilde{L}_i$ are energy and locomotion clipped to the range $[0.1, 1]$, $A_i$ is the acceleration, and $u_i \in \{1, \sqrt{2} \}$ is a heading dependent factor. It ensures that, out of the eight possible directions to move, the four longer diagonal directions have higher cost.
The clipping introduces a minimum cost per step of $0.1$ as well as avoiding division by zero, with the particular choice giving a range $c_i \in [0.1, 11]$ over two orders of magnitude.
The base of the cost is the energy, and the reciprocal of the locomotion introduces a nonlinear penalty to regions of low locomotion.
The result for four different paths are presented in Fig.~\ref{fig:route_optimization}.
Headings and regions with high costs are effectively avoided, as seen by the maximum cost and minimum locomotion in Table~\ref{table:path_plan}.
In general, the paths are navigable, with one notable exception.
In the bottom right corner, path 4 surpasses a region of high costs to the south-west, by repeatedly taking small steps in the north-west and south direction.
This time inefficient part of the path is obviously not realizable due to the turning radius of the vehicle.

\begin{figure}[hbt!]
  \centering
  \includegraphics[width=0.4\textwidth]{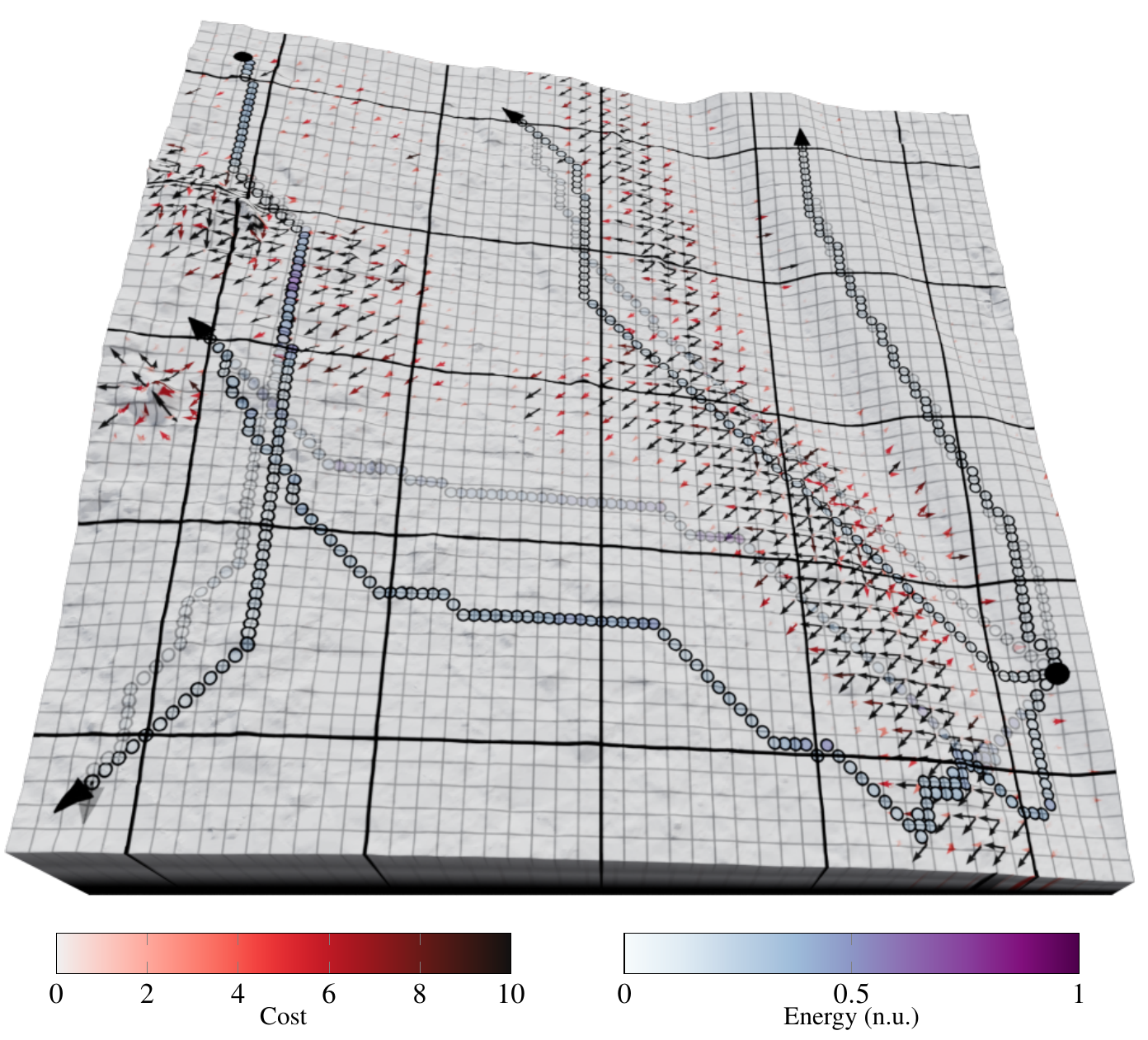}
  \caption{
Comparison of paths using the $c_i$ (semi-transparent) and $c_i^{\text{loc}}$ (opaque) costs, with superimposed vector fields showing the $c_i^{\text{loc}}$ cost in 8 directions.
}
  \label{fig:cost_field}
\end{figure}

We compare the result of the multivalued cost $c_i$, to a cost that only depends on locomotion, $c_i^{\text{loc}} = (1/\tilde{L}_i) u_i$, see Fig.~\ref{fig:cost_field}.
Using the $c_i$ objective results in paths with up to 5\% lower energy consumption, but requires more time to reach the final destination, as seen in Table~\ref{table:path_plan}.
This is natural since using $c_i^{\text{loc}}$ is closely related to optimizing for time.
With the available traversability measures, the cost can be shaped to generate paths efficient in e.g. time or energy, according to preference.
For equipment sensitive to mechanical wear, the cost would be weighted towards acceleration.
This simple method demonstrates the use of traversability maps for path planning, but it is clear that a more sophisticated approach is needed for practical applications.
\begin{table}[ht!]
  \centering
  \begin{tabular}{lcccc} 
    \hline
    Path \#: &  1 & 2 & 3 & 4 \\
    \hline
    Max. cost        & $0.25$ & $0.61$ & $0.89$ & $0.90$  \\
    Min. locomotion  & $0.98$ & $0.79$ & $0.70$ & $0.70$ \\
    Energy c.f. $c_i^{\text{loc}}$  & $100\%$ & $95\%$ & $99\%$ & $95\%$ \\
    Time c.f. $c_i^{\text{loc}}$ & $100\%$ & $103\%$ & $101\%$ & $119\%$ \\
    \hline
  \end{tabular}
  \caption{Properties of planned paths.}
  \label{table:path_plan}
\end{table}

\section{Conclusion}
We conclude that continuous measures of traversability are useful for terrain analysis and planning.
For a given vehicle, simulations with multibody dynamics and generated terrains can be used to train a model that predicts traversability with 90\% accuracy on terrains scanned with at least 0.25 m resolution.
A deep neural network provides the flexibility to couple high-dimensional terrain features with vehicle dynamics.
The trained model depends on the vehicle heading, target velocity, and on detailed features in the topography that a model based only on local slope and roughness cannot capture.
When topography information is available, traversability and cost maps can be generated over large regions at a feasible computational cost.
The maps can be used to produce optimal paths with desired compromise between time, energy, and mechanical wear.
To handle different vehicle configurations, sizes, and loads is straightforward.
Interesting extension are more sophisticated methods for path planning, and to incorporate models of finite soil strength when these become available at high resolution.


\section*{Acknowledgements}
This work was supported by Mistra Digital Forest (Grant 2017/14 \#6) and Algoryx Simulation AB. The simulations were performed on resources provided by the Swedish National Infrastructure for Computing (SNIC dnr 2021/5-234) at High Performance Computing Center North (HPC2N).
The Bo Rydin’s foundation for scientific research (award no F19/17) was co-financing the acquisition of laser scanning data. 

\section*{Appendix A: Supplementary material}
Supplementary videos associated with this article can be
found at \href{http://umit.cs.umu.se/traversability/}{http://umit.cs.umu.se/traversability/}.

\def\cprime{$'$}


\begin{thebibliography}{15}
\expandafter\ifx\csname natexlab\endcsname\relax\def\natexlab#1{#1}\fi
\providecommand{\url}[1]{\texttt{#1}}
\providecommand{\href}[2]{#2}
\providecommand{\path}[1]{#1}
\providecommand{\DOIprefix}{doi:}
\providecommand{\ArXivprefix}{arXiv:}
\providecommand{\URLprefix}{URL: }
\providecommand{\Pubmedprefix}{pmid:}
\providecommand{\doi}[1]{\href{http://dx.doi.org/#1}{\path{#1}}}
\providecommand{\Pubmed}[1]{\href{pmid:#1}{\path{#1}}}
\providecommand{\bibinfo}[2]{#2}
\ifx\xfnm\relax \def\xfnm[#1]{\unskip,\space#1}\fi
\bibitem[{Papadakis(2013)}]{Papadakis2013}
\bibinfo{author}{P.~Papadakis},
\newblock \bibinfo{title}{Terrain traversability analysis methods for unmanned
  ground vehicles: A survey},
\newblock \bibinfo{journal}{Engineering Applications of Artificial
  Intelligence} \bibinfo{volume}{26} (\bibinfo{year}{2013})
  \bibinfo{pages}{1373--1385}.
\bibitem[{Guastella and Muscato(2021)}]{Guastella2021}
\bibinfo{author}{D.~C. Guastella}, \bibinfo{author}{G.~Muscato},
\newblock \bibinfo{title}{Learning-based methods of perception and navigation
  for ground vehicles in unstructured environments: a review},
\newblock \bibinfo{journal}{Sensors} \bibinfo{volume}{21}
  (\bibinfo{year}{2021}) \bibinfo{pages}{73}.
\bibitem[{Chavez-Garcia et~al.(2018)Chavez-Garcia, Guzzi, Gambardella, and
  Giusti}]{Chavez-Garcia2018}
\bibinfo{author}{R.~O. Chavez-Garcia}, \bibinfo{author}{J.~Guzzi},
  \bibinfo{author}{L.~M. Gambardella}, \bibinfo{author}{A.~Giusti},
\newblock \bibinfo{title}{Learning ground traversability from simulations},
\newblock \bibinfo{journal}{IEEE robotics and automation letters}
  \bibinfo{volume}{3} (\bibinfo{year}{2018}) \bibinfo{pages}{1695--1702}.
\bibitem[{Brooks and Iagnemma(2007)}]{Brooks2007}
\bibinfo{author}{C.~A. Brooks}, \bibinfo{author}{K.~D. Iagnemma},
\newblock \bibinfo{title}{Self-supervised classification for planetary rover
  terrain sensing},
\newblock in: \bibinfo{booktitle}{2007 IEEE Aerospace Conference},
  \bibinfo{year}{2007}, pp. \bibinfo{pages}{1--9}.
  \DOIprefix\doi{10.1109/AERO.2007.352693}.
\bibitem[{Bouguelia et~al.(2017)Bouguelia, Gonzalez, Iagnemma, and
  Byttner}]{Bouguelia2017}
\bibinfo{author}{M.-R. Bouguelia}, \bibinfo{author}{R.~Gonzalez},
  \bibinfo{author}{K.~Iagnemma}, \bibinfo{author}{S.~Byttner},
\newblock \bibinfo{title}{Unsupervised classification of slip events for
  planetary exploration rovers},
\newblock \bibinfo{journal}{Journal of Terramechanics} \bibinfo{volume}{73}
  (\bibinfo{year}{2017}) \bibinfo{pages}{95--106}.
  \bibinfo{note}{Manned/Unmanned Ground Vehicles: Off-Road Dynamics and
  Mobility}.
\bibitem[{Quann et~al.(2020)Quann, Ojeda, Smith, Rizzo, Castanier, and
  Barton}]{Quann2020}
\bibinfo{author}{M.~Quann}, \bibinfo{author}{L.~Ojeda},
  \bibinfo{author}{W.~Smith}, \bibinfo{author}{D.~Rizzo},
  \bibinfo{author}{M.~Castanier}, \bibinfo{author}{K.~Barton},
\newblock \bibinfo{title}{Off-road ground robot path energy cost prediction
  through probabilistic spatial mapping},
\newblock \bibinfo{journal}{Journal of Field Robotics} \bibinfo{volume}{37}
  (\bibinfo{year}{2020}) \bibinfo{pages}{421--439}.
\bibitem[{Zhu et~al.(2020)Zhu, Li, Sun, Xu, and Zhao}]{Zhu2020}
\bibinfo{author}{Z.~Zhu}, \bibinfo{author}{N.~Li}, \bibinfo{author}{R.~Sun},
  \bibinfo{author}{D.~Xu}, \bibinfo{author}{H.~Zhao},
\newblock \bibinfo{title}{Off-road autonomous vehicles traversability analysis
  and trajectory planning based on deep inverse reinforcement learning},
\newblock in: \bibinfo{booktitle}{2020 IEEE Intelligent Vehicles Symposium
  (IV)}, \bibinfo{year}{2020}, pp. \bibinfo{pages}{971--977}.
  \DOIprefix\doi{10.1109/IV47402.2020.9304721}.
\bibitem[{Arena et~al.(2021)Arena, Patanè, and Taffara}]{Arena2021}
\bibinfo{author}{P.~Arena}, \bibinfo{author}{L.~Patanè},
  \bibinfo{author}{S.~Taffara},
\newblock \bibinfo{title}{Learning risk-mediated traversability maps in
  unstructured terrains navigation through robot-oriented models},
\newblock \bibinfo{journal}{Information Sciences} \bibinfo{volume}{576}
  (\bibinfo{year}{2021}) \bibinfo{pages}{1--23}.
\bibitem[{Eriksson and Lindroos(2014)}]{Eriksson2014}
\bibinfo{author}{M.~Eriksson}, \bibinfo{author}{O.~Lindroos},
\newblock \bibinfo{title}{Productivity of harvesters and forwarders in ctl
  operations in northern sweden based on large follow-up datasets},
\newblock \bibinfo{journal}{International Journal of Forest Engineering}
  \bibinfo{volume}{25} (\bibinfo{year}{2014}) \bibinfo{pages}{179--200}.
\bibitem[{Suvinen et~al.(2009)Suvinen, Tokola, and Saarilahti}]{suvinen2009}
\bibinfo{author}{A.~Suvinen}, \bibinfo{author}{T.~Tokola},
  \bibinfo{author}{M.~Saarilahti},
\newblock \bibinfo{title}{Terrain trafficability prediction with gis analysis},
\newblock \bibinfo{journal}{Forest Science} \bibinfo{volume}{55}
  (\bibinfo{year}{2009}) \bibinfo{pages}{433--442}.
\bibitem[{Flisberg et~al.(2020)Flisberg, Rönnqvist, Willén, Frisk, and
  Friberg}]{Flisberg2020}
\bibinfo{author}{P.~Flisberg}, \bibinfo{author}{M.~Rönnqvist},
  \bibinfo{author}{E.~Willén}, \bibinfo{author}{M.~Frisk},
  \bibinfo{author}{G.~Friberg},
\newblock \bibinfo{title}{Spatial optimization of ground-based primary
  extraction routes using the bestway decision support system},
\newblock \bibinfo{journal}{Can. J. For. Res.} \bibinfo{volume}{51}
  (\bibinfo{year}{2020}) \bibinfo{pages}{675--691}.
\bibitem[{Perlin(1985)}]{perlin1985image}
\bibinfo{author}{K.~Perlin},
\newblock \bibinfo{title}{An image synthesizer},
\newblock \bibinfo{journal}{ACM Siggraph Computer Graphics}
  \bibinfo{volume}{19} (\bibinfo{year}{1985}) \bibinfo{pages}{287--296}.
\bibitem[{Axelsson(1999)}]{Axelsson1999}
\bibinfo{author}{P.~Axelsson},
\newblock \bibinfo{title}{Processing of laser scanner data—algorithms and
  applications},
\newblock \bibinfo{journal}{ISPRS Journal of Photogrammetry and Remote Sensing}
  \bibinfo{volume}{54} (\bibinfo{year}{1999}) \bibinfo{pages}{138--147}.
\bibitem[{Elmqvist et~al.(2001)Elmqvist, Jungert, Lantz, Persson, and
  Soderman}]{elmqvist2001}
\bibinfo{author}{M.~Elmqvist}, \bibinfo{author}{E.~Jungert},
  \bibinfo{author}{F.~Lantz}, \bibinfo{author}{A.~Persson},
  \bibinfo{author}{U.~Soderman},
\newblock \bibinfo{title}{Terrain modelling and analysis using laser scanner
  data},
\newblock \bibinfo{journal}{International Archives of Photogrammetry Remote
  Sensing and Spatial Information Sciences} \bibinfo{volume}{34}
  (\bibinfo{year}{2001}) \bibinfo{pages}{219--226}.
\bibitem[{{Algoryx Simulations}(2021)}]{AGX2021}
\bibinfo{author}{{Algoryx Simulations}}, \bibinfo{title}{{AGX} {Dynamics}},
  \bibinfo{year}{2021}. \URLprefix
  \url{https://www.algoryx.se/products/agx-dynamics/}.

\end{thebibliography}
\end{document}